\documentclass[11pt]{article}
\usepackage{ijcai17}
\usepackage{times}
\usepackage{url}
\usepackage{latexsym}
\usepackage{graphicx}
\usepackage{enumitem}
\usepackage{comment}
\usepackage{multirow}
\usepackage{booktabs}
\usepackage{xcolor}
\usepackage{amsmath}
\usepackage{bbm}
\usepackage{amsfonts}
\title{Learning Conversational Systems that Interleave Task and Non-Task Content}

\author{Zhou Yu\\
  Carnegie Mellon University \\
  5000 Forbes Avenue, \\
  Pittsburgh, PA, 15213 \\
  {zhouyu@cs.cmu.edu} \\\And
  Alan W Black\\
  Carnegie Mellon University \\
  5000 Forbes Avenue, \\
  Pittsburgh, PA, 15213 \\
  {awb@cs.cmu.edu} \\\And
  Alexander I. Rudnicky \\
  Carnegie Mellon University \\
  5000 Forbes Avenue, \\
  Pittsburgh, PA, 15213 \\
  {air@cs.cmu.edu} \\}

\date{}

\begin{document}

\maketitle
\begin{abstract}
Task-oriented dialog systems have been applied in various tasks, such as automated personal assistants, customer service providers and tutors. These systems work well when users have clear and explicit intentions that are well-aligned to the systems' capabilities. However, they fail if users intentions are not explicit.
To address this shortcoming, we propose a framework to interleave non-task content (i.e.~everyday social conversation) into task conversations. When the task content fails, the system can still keep the user engaged with the non-task content. We trained a policy using reinforcement learning algorithms to promote  long-turn conversation coherence and consistency, so that the system can have smooth transitions between task and non-task content.
To test the effectiveness of the proposed framework, we developed a movie promotion dialog system. Experiments with human users indicate that a system that interleaves social and task content achieves a better task success rate and is also rated as more engaging compared to a pure task-oriented system.
%\dk{Out of curiosity, is there a reason you didn't just incorporate your stuff into an established task-oriented dialog system, instead of creating the movie one?}

%This framework is especially useful for situations where users lack a clear calibrated goal with the system.
\end{abstract}

\section{Introduction}
The most familiar communication setting for people is talking to another human. Therefore, normal users would transfer their human-human behavior patterns and expectations to interactions with a system. For example, though users quickly learned that Microsoft Cortana (a personal assistant) could not handle social content, 30\% of the total user utterances addressing it are social content \cite{jiang2015automatic}. Therefore, one possible way to improve conversational system performance is to imitate human behaviors. Countless observations suggest that human conversations usually interleave social content with task content \cite{schegloff1968sequencing}. For example, we usually open a conversation with ``How are you doing?"; we also divert to social topics during meetings; and we would most likely end our workday conversations with chitchat of our weekend plans. 
However, traditional conversational systems are mainly task-oriented. These systems can complete tasks such as booking airline tickets \cite{zue1994pegasus}, searching for bus information \cite{raux2005let}, and making restaurant reservations \cite{jurcicek2011real}. These systems do not involve social content mainly because the task is relatively simple and the user intention and system capability are well calibrated. 

To move our current systems to tackle tasks that are more complex and especially those where most users do not have clear intentions, we propose a dialog framework to fuse task and non-task conversation content. To achieve the content transition smoothness, we trained dialog policies using reinforcement learning algorithms. We built an example dialog system, a movie promotion system that promotes a specific movie according to users' interests and uses social conversation to engage users to complete the task. There are several types of audience research conducted by film distributors in connection with domestic theatrical releases \cite{martin1982}. Such audience research can cost \$1 million per movie, especially when scores of TV advertisements are tested and re-tested. Therefore, we argue that having conversational system to elicit audience information voluntarily to replace paid surveys would reduce the cost and improve the survey quality.

We published the source code of the software implementation of the framework, an example movie promotion system, and the conversation data collected with human users.\footnote{\url{https://github.com/echoyuzhou/ticktock_text_api}}
The framework is general and applicable in different domains, such as political surveying, language learning, and public health education. The theoretical framework and the software implementation enables researchers and developers to build example dialog systems in different domains, hopefully leading to big impact beyond discourse and dialog research.

\section{Related Work}
Current task-oriented dialog systems focus on completing a task together with the user. They can perform bus information search \cite{raux2005let}, flight booking \cite{zue1994pegasus}, direction giving \cite{yu2015sigdial}, etc. However, these systems can only focus on one task at a time. The famous personal assistants, such as Apple's Siri are composed of many of these single-task systems. These single-task systems' underlying mechanisms are mainly frame-based or agenda-based \cite{rudnicky1999agenda}. %\dk{I don't know what these are, but I'm not sure that's that important that I know} %The architecture is a set of pre-defined frames link to user intent, such as booking a flight. Each frame is composed of one or several semantic slots that linking to templates. Templates are then used to elicit task information. 
The architecture of traditional dialog systems is slot-filling, which pre-defines the structure of a dialog state as a set of slots to be filled during the conversation. For an airline booking system, an example slot is ``destination city". An example corresponding system utterance generated from that slot is "Which city are you flying to?" Recently researchers have also started to look into end-to-end learning for task-oriented systems. Though the progress is still preliminary \cite{bordes2016learning}, the premise of having a learning method that generalizes across domains is appealing. 

Differing from task-oriented systems, non-task-oriented systems do not have a stated goal to work towards. Nevertheless, they are useful for social relationship bonding and have many other use cases, such as keeping elderly people company \cite{higashinaka2014towards}, facilitating language learning \cite{jia2008use}, and simply entertaining users \cite{yu2016sigdialeng}. Because non-task systems do not have a goal, they do not have a set of restricted states or slots to follow. A variety of methods were therefore proposed to generate responses for them, such as machine translation \cite{ritter2011data}, retrieval-based response selection \cite{banchs2012iris}, and sequence-to-sequence models with different structures, such as, vanilla recurrent neural networks \cite{google}, hierarchical neural models \cite{serban2015building}, and memory neural networks \cite{dodge2015evaluating}.

However, there is no research on combining these two types of dialog systems so far. Therefore, our work is the first attempt to create a framework that combines these two types of conversations in a natural and smooth manner for the purpose of improving conversation task success and user engagement. Such a framework is especially useful to handle users who do not have explicit intentions.

To combine these two types of conversation systems smoothly, we trained a response selection policy with reinforcement learning algorithms. Reinforcement learning algorithms have been used in traditional task-oriented systems to track dialog states \cite{williams2007partially}. They have also been used in non-task oriented systems. The Q-learning method was used to choose among a set of statistical templates and several neural model generated responses in \cite{yu2016sigdial}, while the policy gradient method was used in \cite{li2016deep}. Different from these pure task or pure non-task systems, we applied reinforcement learning algorithms to train policies that choose among task and non-task candidate responses to optimize towards a coherent, consistent and informative conversation with respect to different users.

\section{Framework Description}
\begin{figure}
\centering
\includegraphics[scale=.35]{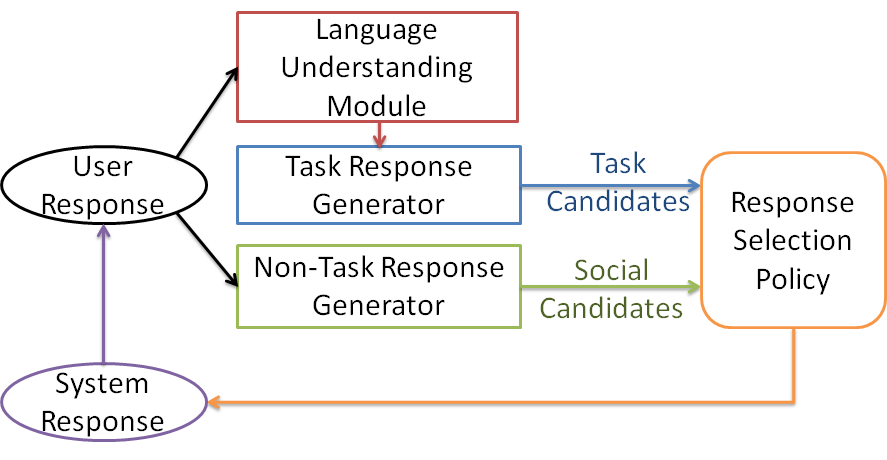}
\caption{Framework Architecture. \small{A user utterance is sent to both a language understanding module and a non-task response generator. The understanding model then extracts useful information to help a task response generator to produce task-oriented candidates. Simultaneously, the non-task response generator also produces non-task candidates. Finally, a response selection policy selects among all the candidates to produce a system response.} } % it is not sidecaption, how to fix it.
\label{fig:flow}       
% \vspace{-5mm}
\end{figure}
The framework has four major components: a \textit{language understanding module}, a \textit{task response generator}, a \textit{non-task response generator} and a \textit{response selection policy}. Figure~\ref{fig:flow} shows the information flow among these components. A user utterance is sent to both the language understanding module and the non-task response generator. The understanding model then extracts useful information to help the task response generator to produce task-oriented candidates. Simultaneously, the non-task response generator also produces non-task candidates. Finally, the \textit{response selection policy} selects among all the candidates to produce a system response. We will discuss each component in details below. %\dk{This is pretty much the same text as the caption, I'm not sure you need the caption if you're going to repeat the same thing.}

The language understanding module extracts information required by the task from user responses, and sends the information to the task response generator. There are many approaches to perform language understanding. A rule-based keyword-matching method is sufficient for simple tasks \cite{he1998language}, while learning-based methods perform better in complex tasks. The understanding task can be formulated as different learning problems: for example a supervised classification problem that can be solved by algorithms, such as Support Vector Machines \cite{williams2015fast}; or a sequence labeling problem that can be solved by algorithms, such as recurrent neural networks \cite{mesnil2013investigation}. 

The task response generator produces a list of task response candidates considering the information passed from the language understanding module. The most common method to generate these candidates is pre-defined statistical templates that are used in traditional slot-filling task oriented systems \cite{rambow2001natural}. Each template intends to elicit information required by the task. While the non-task response generator produces a list of social response candidates. This component could be implemented by various methods mentioned in the related work, such as machine translation, sequence-to-sequence, keyword retrieval, or an ensemble of these methods. The non-task content intends to make the conversation more coherent and engaging, thus keeping users in the conversation. Therefore, users would have more chance to complete the task.

The response selection policy sequentially chooses among all the response candidates (non-task or task candidates) to optimize towards natural and engaging interactions. Different reinforcement learning algorithms could be applied to train the policy, such as Q-learning \cite{sutton1998reinforcement}, SARSA \cite{sutton1998reinforcement}, and policy gradient \cite{sutton1998reinforcement}. Different reinforcement learning algorithms would be preferred with respect to other framework components' implementation. For instance, if the number of response candidates are limited and the dialog states can be represented in tabular forms, then Q-learning is sufficient. While if the state representation is continuous, then the policy gradient method is preferred. In the next section, we will use a movie promotion task as an example to describe example algorithms in details.
% 
%1)Should talk about limitations, which is the social system trained, need to have existing data that is on the similar topics, otherwise it won't work well. 
%We also need to give stats about the different individual metrics of the social systems to validate how good our social system is. The other thing is 

\section{Example Dialog Task: Movie Promotion}
The movie promotion task has a goal to promote a movie taking account of users' preferences. To imitate human-human conversation, the framework suggests to open conversations with a non-task conversation topic related to the task. Therefore, the movie promotion system starts a conversation saying: ``Hello, I really like movies. How about we talk about movies?" We then describe the implementation details of each component.

\subsection{Language Understanding Module}
Because the user responses' sentence structure is relatively simple and the information to extracted is mainly `yes/no' and named entities, we used a shallow parser to provide features, and then a set of pre-designed rules for each task template. For example, for the response to the ``If Seen Movie?" template, we simply used a key-word matching rule to classify the utterance to `yes' or `no' categories. The extracted information is then fed to the task response generator. The collected information is also useful for task analysis. %\dk{I'm a bit confused, why are you assuming that the user responses will be simple?}

\subsection{The Task Response Generator}
The task response generator produces response candidates using a sets of pre-defined language generation templates considering the information received from the language understanding module. We designed the following eight templates to approach the goal of promoting a movie, such as ``Captain America: Civil War". 

\begin{itemize}
    
\item \textbf{Elicit movie type}: The system elicits the user's preferred movie type, e.g. ``Do you like superhero movies or Disney movies?". %\dk{Minor issue here, most superhero movies are Disney movies, so you might want to reword this (Marvel is owned by Disney).}

\item \textbf{Introduce favorite superhero}: The system expresses its favorite superhero is Captain America, in order to lead to the movie for promotion, e.g. ``My favorite superhero is Captain America."
 
\item \textbf{Ground on superhero}: We crawled wiki-webpages to build a superhero knowledge database that includes all the superheroes' details, such as real name, eye colors, origin, etc. If the user mentions any superhero, the system will talk about some attributes of that superhero, e.g.  ``I really like Iron Man's blue eyes." 

\item \textbf{Discuss relevant movie}: The system talks about a relevant movie the user mentioned before, e.g. ``I really like the first Avenger movie, have you seen it before?" 

\item \textbf{Discuss movie detail}: The system further elaborates on the details of the mentioned relevant movie, e.g.``I really liked the first Avenger movie. When Iron Man came back alive, I cried." 

\item \textbf{Saw the movie}: The system asks the user if they saw the movie for promotion, e.g.``Have you seen the new superhero movie, `Captain America: Civil War'?" 

\item \textbf{Promote the movie}: The system promotes the intended movie. e.g. ``One of my friends just saw `Captain America: Civil War'. He told me it is a really nice movie, much better than the previous Captain America movie."

\item \textbf{Invite to the movie}: The system suggests to see the promoted movie together, e.g. ``Do you want to see Captain America: Civil War together?" %\dk{Wait I don't unerstand, what happens when someone wants to see the movie with the program?}

\end{itemize}

\subsection{Non-Task Response Generator}
The non-task generator provides three types of candidate responses via three methods described below. 
\begin{itemize}
\item A keyword retrieval method trained on a CNN interview corpus \cite{yu2015ticktock}.    
\item A skip-thought vector model \cite{kiros2015skip} trained on the Movie Subtitle dataset \cite{lison2016opensubtitles2016} 
\item A set of conversation strategies that generated via statistical templates that emulate human-human conversation strategies, in order to foster user coordination, understanding and adaptation. In particular, we designed three types of conversation strategies following \cite{yu2016sigdial}. 
\begin{itemize}
\item \textit{Active participation strategies} engage users by actively contributing to the conversation, such as asking more information on the current topic \cite{wendler}. 
\item \textit{Grounding strategies} assist open-domain natural language understanding. Grounding strategies were automatically synthesized via leveraging knowledge-base (e.g. Google Knowledge Graph) information and natural language processing algorithms, such as named-entity detection and statistical language generation. %For instance, if an ambiguous named-entity ``Clinton" is mentioned, the system would provide users with details of two popular  ``Clintons" from the knowledge base to collaboratively resolve the ambiguity with users by asking:  ``Are you talking about Bill Clinton or Hillary Clinton? " %\dk{I don't really understand this one}
\item \textit{Personalized strategies} support adaptions to users by leveraging automatically extracted information from individual user's conversation history. An example personalized strategy is to suggest talking more about a certain topic knowing that the user was previously engaged in that topic.    
\end{itemize}

\end{itemize}

\subsection{Response Selection Policy}
The response selection policy is designed to select candidates provided by the two response generators. Conversation processes can be modeled as Markov Decision Processes (MDPs) \cite{williams2007partially}. Therefore, we used reinforcement learning algorithms to train a policy that optimizes towards conversations with long-term coherence, consistency, variety, and continuity. %\dk{This is a strange sentence, I'm not sure what you're trying to say here as the second part of the sentence seems unrelated to the first part}
\begin{comment}
\begin{figure*}[htbp]
\begin{equation}
Q_{t+1}(s_{t},a_{t}) \leftarrow {Q_t(s_t,a_t)} + \alpha_t(s_t,a_t) \cdot \left( R_{t+1} + \gamma \max_{a}Q_t(s_{t+1}, a) - Q_t(s_t,a_t) \right)
\end{equation}
\end{figure*}
\end{comment}
Specifically, we used Q-learning, a model-free reinforcement learning algorithm. Because it handles discrete states well and learns a Q table that supports a model that makes both debugging and interpretation easier. Another advantage of Q-learning is that it also makes encoding expert knowledge easier, as it is a model-free algorithm. By encoding expert knowledge, the search space can be reduced, thus making the algorithm converge faster. In all, Q-learning is a good choice with respect to the implementation choice of other components of the system.

In a reinforcement learning setting, we formulate the problem as $(\mathbb{S},\mathbb{A},\mathbb{R},\gamma,\alpha)$, where $\mathbb{S}$ is a set of states that represents the system's environment, in this case the conversation history so far. $\mathbb{A}$ is a set of actions available per state. In our setting, the actions are strategies available. By performing an action, the agent can move from one state to another. Executing an action in a specific state provides the agent with a reward (a numerical score), $\mathbb{R}(s,a)$. The goal of the agent is to maximize its total reward. It does this by learning which action is optimal to take for each state. The action that is optimal for each state is the action that has the highest long-term reward. This reward is a weighted sum of the expected values of the rewards of all future steps starting from the current state, where the discount factor $\gamma \in (0, 1)$ trades off the importance of sooner versus later rewards. $\gamma$  may also be interpreted as the likelihood to succeed (or survive) at every step. The algorithm therefore has a function that calculates the quantity of a state-action combination, $\mathbb{Q}: \mathbb{S} \times \mathbb{A} \rightarrow \mathbb{R}$. The core of the algorithm is a simple value iteration update. It assumes the old value and makes a correction based on the new information at each time step, \textit{t}. %See Equation (1) for details of the iteration function.
The critical part of the modeling is to design appropriate states, actions and a corresponding reward function.\\
\\
\noindent\textbf{State and Action Design}
It is difficult to design states for conversational systems, as one slight change from a user response may lead to a completely different conversation. We reduced the state space by incorporating extra knowledge, the statistics obtained from conversational data analysis. Following \cite{yu2016sigdial}, we include features: turn index, number of times each strategy executed, sentiment polarity of all previous utterances, coherence confidence of the response, and most recently used strategy. We constructed the reward table based on the statistics provided in \cite{yu2016sigdial} %\dk{What previous experiment? Is this from the paper you cited?}
. We utilized expert knowledge to construct rules to constrain the reward table. For example, if certain strategies have been used before, then the reward of using it again immediately is heavily penalized. Please see a detailed list of constrains in the appendix. These constraints may result in less optimal solutions, but reduce the state and action search space considerably. The actions are simply all the response candidates produced by both task and non-task generators.\\
\\
\noindent\textbf{Reward Function Design}
We designed the reward function to be a linear combination of four metrics: \textit{turn-level appropriateness (App)}, \textit{conversational depth (ConvDepth)}, \textit{information gain (InfoGain)}, and \textit{conversation length (ConvLen)}. Except for the first metric which is an immediate reward, all the others are delayed rewards. We first initiated weights associate to each metric manually, and refine them in training later. One difficulty for interactive tasks is that during training, we can not interrupt the interaction flow by asking users to give turn-by-turn immediate feedback. To solve this problem, we used pre-trained predictors to approximate these metrics, which is similar to inverse reinforcement learning. We will describe all the metrics along with methods to approximate them in the details below. 

\begin{itemize}

\item \textbf{Turn-level appropriateness (App)}: reflects the coherence of the system's response in each conversational turn. For later experiments, we adopted the same annotation scheme in \cite{lrec} and used the appropriateness predictor provided in \cite{yu2016sigdial}, which trained on 1256 annotated turns. %using a v-Support Vector \cite{chang2011} with a RBF Kernel. 
The performance of the automatic appropriateness detector is 0.73 in accuracy (majority vote is 0.5 in accuracy). 

\item \textbf{Conversation depth (ConvDepth)}: reflects the number of consecutive utterances that share the same topic. We followed the same annotation scheme in \cite{yu2016sigdial}: labeling conversations having ten or more consecutive turns to be deep and others as shallow. We used the predictor trained on 100 conversations from \cite{yu2016sigdial}. It has 72.7\% in accuracy, while the majority vote baseline accuracy is 63.6\%. 

\item\textbf{Information gain (InfoGain)}: reflects the number of unique words that are introduced into the conversation from both the system and the user. We believe that the more information the conversation has, the better the conversational quality is. This metric is calculated automatically by counting the number of unique words after the utterance is tokenized. %\dk{Wouldn't it be good to ignore things like prepositions, pronouns, etc? I feel like really nouns are the only that really matter even though I'm not really a linguist haha}

\item\textbf{Conversation Length (ConvLen)}: reflects how long the user want to stay in the conversation. We approximated it by the number of turns in the overall conversation. The assumption is that the more the users want to interact with the system, the better the system is.
\end{itemize}

\section{Experiments}

We built three systems for the movie promotion task.
\begin{itemize}
    \item \textbf{\textit{Task-Global}}: This does not have a non-task response generator and therefore can only output task responses. Its response selection policy is trained with the entire conversation history.
    \item \textbf{\textit{Mix-Local}}: This has both the task and non-task generators, and a response selection policy considers three previous turns as interaction history. 
    \item \textbf{\textit{Mix-Global}}: This has both the task and non-task generators, and a response selection policy trained with the entire interaction history.
\end{itemize}

We used another chatbot, A.L.I.C.E.\footnote{http://alice.pandorabots.com/}, which is powered by rules, as a user simulator to train the response selection policy for all systems. During training, we restart conversations if the user simulator repeats the same utterance. It took 200, 1000, and 8000 conversations respectively for the Task-Global, Mix-Local, and Mix-Global systems to converge.

Besides testing the system with the user simulator, we also recruited human users on crowdsourcing platforms (Amazon Mechanical Turk and Crowd Flower). We recruited crowd workers that are located in the U.S. and have a previous task approval rate that is higher than 95\%. We asked them to interact with the system for as long as they want. They were only allowed to interact with one of the three systems once, thus preventing them exploiting the task. Within two days, 150 crowd workers participated and produced 50 conversations for each system. The users also reported their gender and age range after the interaction. In addition, we asked them to rate how engaged they were throughout the conversation in a 1-5 Likert scale (a higher score indicates more engagement). Table~\ref{table:example} shows an example conversation produced by the Mix-Global system. We consider the task to be successful if the user received all the task responses. Apart from the task success rate, the self-reported user engagement score is also included for evaluation. 

\section{Results and Analysis}
From experiments, we found that the system that interleaves non-task content and considers the entire conversation history in policy training performs the best in both simulated and real-world settings. We discuss detailed statistics below:\\

\begin{figure}[htb]
\centering
\graphicspath{ {figures/} }
\includegraphics[scale=.40]{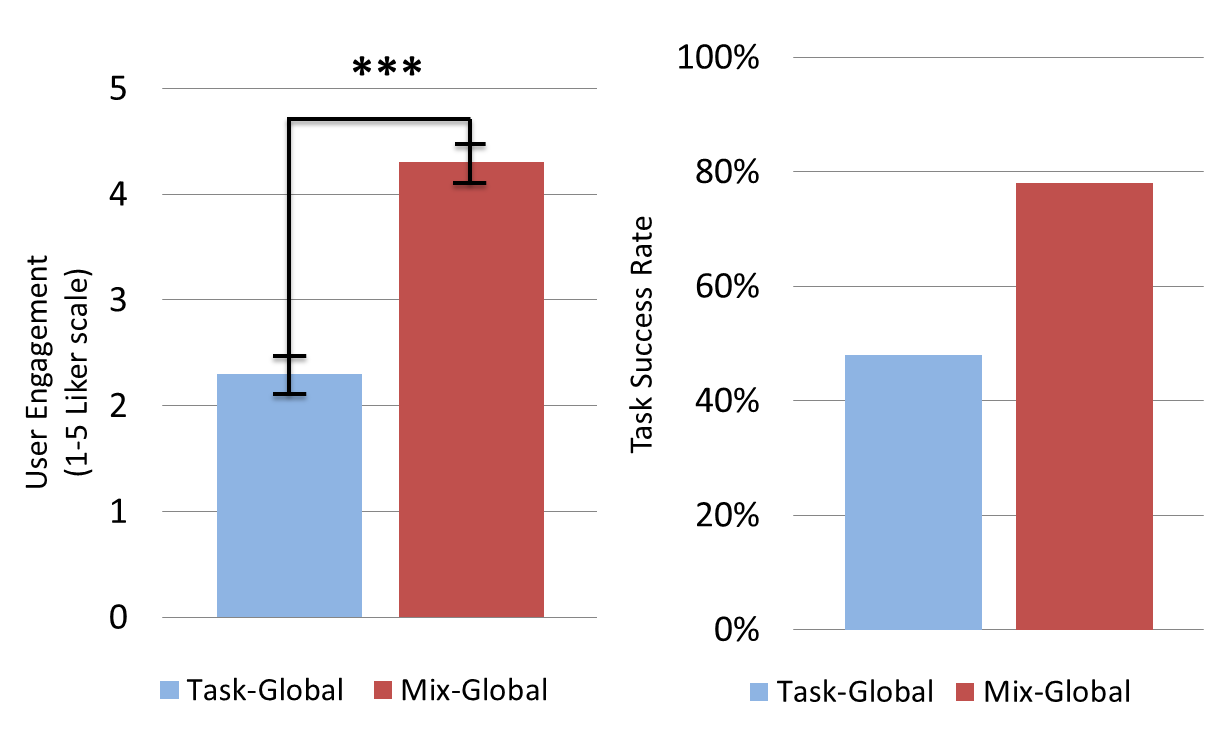}
\caption{The Mix-Global system outperformed the Task-Global system on both user engagement and task success.}
\label{fig:ijcai}       
\end{figure}

\noindent\textbf{Finding 1}: Involving non-task content makes the user more engaged and at the same time increases the task success rate. Figure~\ref{fig:ijcai} shows that the Mix-Global system outperformed the Task-Global system with respect to both the task success rate and the user self-reported engagement rating with statistical significance (t-test, p$<$ 0.001). The average user self-reported engagement is 2.3 (SD=0.42) for the Task-Global system, and 4.4 (SD=0.21) for the Mix-Global system. The task success rate of the Mix-Global system outperformed (78\%) the Task-Global system (23\%) with a big margin. The good performance comes from the fact that in the Mix-Global system, the non-task responses handled the user utterances that task responses could not handle. Therefore the user could remain in the conversation,  providing the system with more opportunities to complete the task.

\begin{table}
%\small{    
\begin{tabular}{p{8.5cm}}
\hline
1 \textit{Non-Task}: Hello, I really like movies. How about we talk about movies? \\
2 \textit{User}: I like watching movies too. \\
3 \textit{Task}: Do you like superhero  movies or Disney movies?\\
4 \textit{User}: I like superhero movies. \\
5 \textit{Task}: My favorite superhero is Captain America.\\ 
6 \textit{User}: I like Spider-man.\\
7 \textit{Non-Task}: Do you watch them with your kids?\\ 
8 \textit{User}: I don't have any children. \\
9 \textit{Non-Task}: What I meant to say was, what is it that you hate?\\
10 \textit{User}: I hated the last Fantastic Four movie.\\
11 \textit{Non-Task}: Are you talking about Fantastic Four, the 2005 film.\\
12 \textit{User}: Yes. I am.\\
12 \textit{Task}: I really like the first Avenger movie, have you seen it before?\\
\hline
\end{tabular}
%}    
\caption{An example conversation between a human user and the Mix-Global system. Utterances 7, 9, and 11 are social responses that selected by the policy to keep the conversation coherent, consistent and informative. }
\label{table:example}
\end{table}
%\dk{You repeat the same thing in the caption here, I'm kinda against this as before, but if this is how you guys do things that's fine} 
Table \ref{table:example} shows an example conversation between the Mix-Global system and a human user. Utterances 7, 9, and 11 are system non-task responses. They were selected by the policy for yielding coherent and consistent conversations. For example, utterance 9 ``What I meant to say was, what is it that you hate?" was chosen to promote topic consistency and at the same time to preserve local coherence. Non-task content also contributes to content variety. Overall, the policy trained with reinforcement learning algorithms makes the transition between non-task and task smoother. We suspect that is the major reason that users find the system that interleaves non-task and task content more engaging %\dk{I'm surprised by this, I would imagine it's the smooth transition. If I was talking to someone and they didn't vary their conversation I wouldn't mind too much, but if they kept trying to push their own agenda of me watching Civil War with them, I'd be more turned off.}
. However, the Mix-Global system still fails to engage some users who do not have any interest in superhero movies. For example, one user replied: ``No, I haven't seen [Captain America: Civil War], I am not a stupid teenager or a stupid robot." and left the conversation early. \\

\begin{figure}[htb]
\centering
\graphicspath{ {figures/} }
\includegraphics[scale=.40]{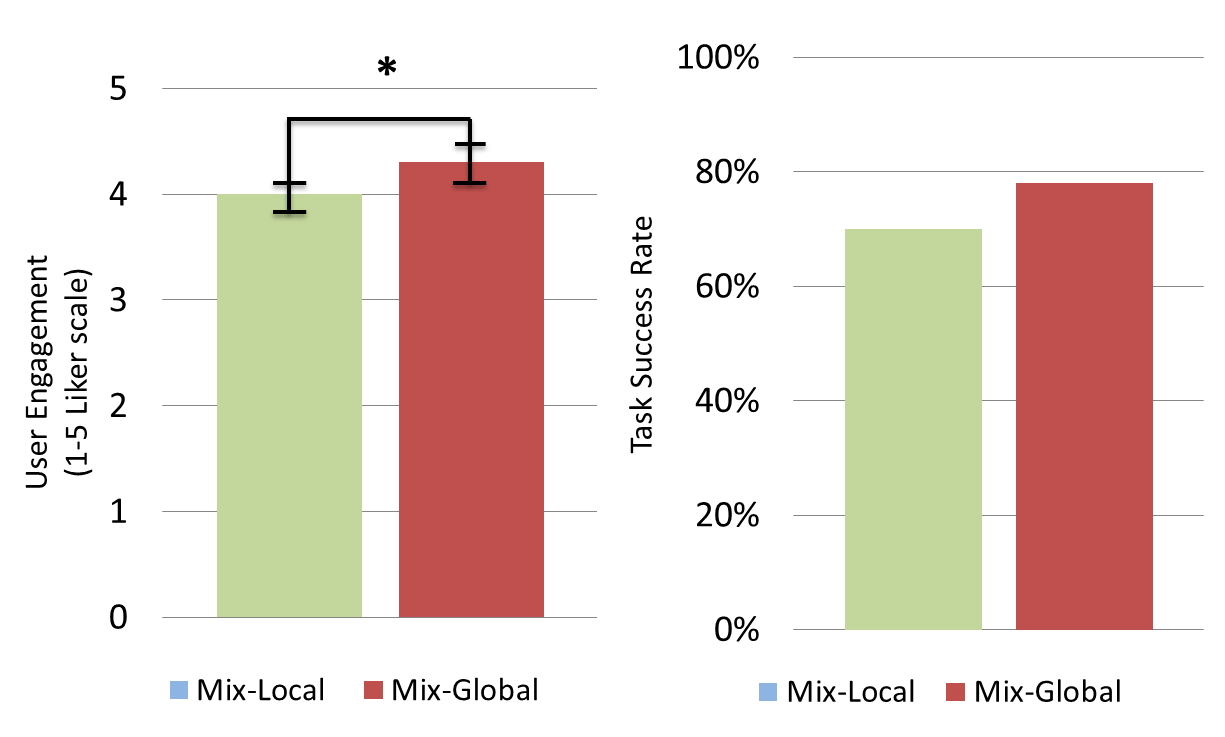}
\caption{The Mix-Global system outperformed the Mix-Local system on both user engagement and task success.}
\label{fig:ijcai2}       
\end{figure}

\noindent\textbf{Finding 2}: Incorporating longer interaction history engages users better and improves task success rate. Figure \ref{fig:ijcai2} shows that the user self-reported engagement of the Mix-Local system (4.0 (SD=0.32)) outperformed the Mix-Global system (4.4 (SD=0.21)) with a moderate statistical significance (p$<$0.05). The Mix-Global system also outperformed the Mix-Local system on task completion rate (70\% VS. 78\%). Therefore, statistics suggest that a system considers the entire conversation history in planning is more engaging and completes the task more often, compared with a system that only takes the previous three utterance history. The small margin is caused by the fact that the task conversation is not very long (8 responses in total). Therefore, we expect a stronger performance impact for tasks that are more complex that require longer conversations.  \\

\noindent\textbf{Finding 3}: We also found that the system interleaves social content and is powered by a policy trained with the entire conversation history performs best in all the individual metrics in the reward function in both simulated and real-world (interacting with human users) settings (see Table 2 and 3 for details). We follow the convention in \cite{yu2016sigdial} to consider conversations that have 10 or more consistent turns on the same topic to be deep conversations. Because the Task-Global system only produces conversations with 8 turns, we do not compare it on the conversation depth metric. Among all the metrics, we found response coherence (App) improved most by incorporation social content, indicating the smooth transition has the biggest impact on the overall performance improvement. However, we also found that all systems' performance is always lower in the real-world setting compared to the simulated setting. Probably because the user simulator is limited and it was powered by an extensive set of rules.  %\dk{What is this about real-world and simulated settings?}

 \begin{table}[htbp]
 \centering
    \begin{tabular}{l|cccc}
    \hline
    System& App & ConvDepth & InfoGain & ConvLen\\
    \hline
    Task-Global & 32.5\% & NA & 34.5 & 5.7\\
    \hline
    Mix-Local & 76.3\% & 77\% & 52.4 & 13.7\\
    \hline
    Mix-Global & 80.1\% & 88\% & 66.8 & 17.4\\
    \hline
    \end{tabular}
    \caption{The Mix-Global system performed the best when interacting with a simulated user.}
    \label{simulate}
\end{table}

\begin{table}[htbp]
    \centering
    \begin{tabular}{l|ccccc}
    \hline
         System&  App & ConvDepth & InfoGain & ConvLen\\
         \hline
         Task-Global &31.3\% & NA & 39.3 & 5.3\\
         \hline
         Mix-Local & 73.0\% & 71\% & 62.4 & 13.0\\
         \hline
         Mix-Global &76.7\% & 79\%  & 67.2 &15.8\\
         \hline
    \end{tabular}
    \caption{The Mix-Global system performed the best when interacting with human users. }
    \label{table: result1}
\end{table} 

 %All the slots are binary questions, so we simply check if the replied sentence contains ``no" and ``n't" in the sentence. If the reply contains these words, the slot is filled with ``Negative", otherwise ``Positive". If in a conversation these strategies have not been asked, then the slot is marked as ``NULL".
\vspace{-2mm}
\section{Movie Promotion Task Data Analysis}
To validate the usability of the example movie promotion system, we conduct a brief analysis on the information collected by the system. We found three interesting suggestive phenomena that would be of interest to the film industry, though the user sample pool is relatively small (150 participants) and biased (crowd workers located in the U.S.). \\
%\vspace{1mm}

\noindent\textbf{Participants prefer superhero movies over Disney movies.} %\dk{Yeah so I mentioned this before but superhero movies are often Disney movies due to Disney's purchase of Marvel, so you might want to reword this.}
We found that 42\% of participants preferred superhero movies, 22\% of participants preferred Disney movies, and 36\% of participants liked both or neither. Overall, superhero movies are more popular than Disney movies among our participants. One possible explanation is that we surveyed only U.S. residents.

\noindent\textbf{Men prefer superhero movies.} 
%Among the 150 participants, 84 of them were men. 
Among the 150 participants, 102 (57 men) of them were asked if seen the promoted movie and 42 (41\%) of them did. Among them 57.9\% of men saw the film, while only 20.0\% of women did. 90 participants were asked if they would like to see the promoted movie, and 80\% of them said yes. Men were more likely to go along with the invitation than women (77.8\% of men and 68.3\% of women said yes). Both findings indicate that men compared to women are more interested in the promoted movie. \\%\dk{You might want to include how many men and women are in the 102 and 42.}

%14 of them saw the movie, in which 11 of them are men. 20 people haven't seen the movie, in which only 8 of them are men.
%Although we had 6 more male participants in the task compared to female participants, we still find superhero movies are more popular among men with statistically significance ($p<0.05$). 
%21 people said yes and among them 14 people are men. 9 people said no, and in which 4 of them are men. 
\noindent\textbf{Twenty-year-olds like superhero movies most.}
We separate the participants into five different age groups: below 20, 20-30, 30-40, 40-50, and 50 above. 
People who had seen the promoted movie spread across all age groups, and mostly concentrate in the 20-30 age group (27 out of 42). Even though our participants are mostly in this age group (78 out of 150), we still found that there are more participants in their 20s who have seen the promoted movie compared to other groups with statistical significance (t-test $p<0.05$). In addition, compared to participants in other age groups, participants in their 20s are more likely to accept the invitation to see the promoted movie together (39 out of 42).
\\

Statistics reported by the Motion Picture Association of America shows a similar trend. They found that females accounted for only 42\% of the audience for ``Captain America: The Winter Soldier", and 18-25 age group attended movie most per capita is the 18-25 in 2015 \cite{theatrical}.

%\dk{I think the main point of this stuff is that you want to say you got good information about movie things automatically (via your bot). In which case, if you show that your findings agree with statistics that movie producers or w/e have collected, it'll show that your info is good and better drive the point home.}

\section{Conclusion and Future Work}
We proposed a framework to develop conversational systems that interleave task content with non-task content in a natural manner, in order to improve conversation task success and user engagement. The framework is mainly powered by a statistical policy that selects among task and non-task candidate responses to optimize towards long-term conversation effectiveness. This framework is general enough to apply to many tasks, and is especially useful for tasks where users often do not have any concrete intentions. We also designed an example system under the framework that promotes movies to validate the framework. Through experiments conducted with both a simulator and real human users, we found the system that interleaves non-task content achieved better task success and user engagement. We also validated the systems usage by finding some interesting phenomena from the data collected that correlates with statistics reported by theatrical associations. Although our participant sample size is small, we believe these statistics could still help movie makers understand their markets better.

While the example task centers on movie promotion, there is nothing specific to the task design which could not be transferred immediately to other domains such as product recommendations, public health education, customer services and political surveys. Future work would further investigate different types of representations for reinforcement learning algorithms so as to make the framework generalize well across domains.

\end{document}